\documentclass[10pt,twocolumn,letterpaper]{article}

\usepackage{iccv}
\usepackage{times}
\usepackage{epsfig}
\usepackage{graphicx}
\usepackage{amsmath}
\usepackage{amssymb}
\usepackage{multirow}
\usepackage{booktabs}
\usepackage{subfigure}
\usepackage[misc]{ifsym}

\usepackage[breaklinks=true,bookmarks=false]{hyperref}

\iccvfinalcopy 

\def\iccvPaperID{****} 

\ificcvfinal\fi

\begin{document}
\newcommand\correspondingauthor{\thanks{Corresponding author.}}
\title{Towards Building More Robust Models with Frequency Bias}

\author{Qingwen Bu\textsuperscript{1,3}  \quad \quad \quad \quad Dong Huang\textsuperscript{2}\correspondingauthor \quad \quad \quad \quad Heming Cui\textsuperscript{1,2}\\
\textsuperscript{1}Shanghai Artificial Intelligence Laboratory \quad  \textsuperscript{2}The University of Hong Kong \\  \textsuperscript{3}Shanghai Jiao Tong University\\
{\tt\small qwbu01@sjtu.edu.cn,\quad u3008427@connect.hku.hk,\quad heming@cs.hku.hk}
}

\maketitle
\ificcvfinal\thispagestyle{empty}\fi

\begin{abstract}
The vulnerability of deep neural networks to adversarial samples has been a major impediment to their broad applications, despite their success in various fields. Recently, some works suggested that adversarially-trained models emphasize the importance of low-frequency information to achieve higher robustness. While several attempts have been made to leverage this frequency characteristic, they have all faced the issue that applying low-pass filters directly to input images leads to irreversible loss of discriminative information and poor generalizability to datasets with distinct frequency features. This paper presents a plug-and-play module called the Frequency Preference Control Module that adaptively reconfigures the low- and high-frequency components of intermediate feature representations, providing better utilization of frequency in robust learning. Empirical studies show that our proposed module can be easily incorporated into any adversarial training framework, further improving model robustness across different architectures and datasets. Additionally, experiments were conducted to examine how the frequency bias of robust models impacts the adversarial training process and its final robustness, revealing interesting insights.
\end{abstract}

\section{Introduction}
As deep learning methods are making a splash in various fields, their security and robustness are drawing more and more attention from both academia and industry. DNNs for image classification have been proven to be easily fooled by adversarial examples, with imperceptible perturbations added to natural images. The vulnerability is also uncovered in many other safety-critical fields like medical diagnosis~\cite{finlayson2019adversarial} and autonomous driving~\cite{deng2020analysis}. Adversarial training is regarded as a trustworthy approach for producing robust models, and considerable subsequent effort has been done to increase its efficiency~\cite{shafahi2019adversarial, wong2020fast} and efficacy~\cite{zhang2020attacks,zhang2020geometry}. Despite continuously advanced robust learning methods, the relationship between some intrinsic characteristics of the model structure, such as frequency characteristics, and the robustness it exhibits seems to be rarely discussed.

The emergence of vision transformers~(ViTs) revealed a novel architecture that performs on par or even better than convolutional networks~(CNNs) in many vision tasks. As more and more in-depth work unfolds, some of ViT's fundamentals and characteristics are exposed. Several recent works try to improve the performance of ViTs with a frequency lens. 
\cite{shao2021adversarial} proposed naturally trained ViTs are more robust to adversarial attack, especially for high-frequency perturbations. It is also noticed that ViTs reduce high-frequency signals while CNNs amplify them~\cite{park2022vision}. \cite{wang2022anti} successfully enhances ViT performance by deliberately retaining high-frequency information. The complementarity of CNNs and ViTs exhibited in the frequency domain is an intriguing starting point for studying and further improving the robustness of the model. But the relations between robustness and the frequency bias of models have not been fully studied and utilized in the context of robust learning.

Research~\cite{geirhos2018imagenet} has suggested that deep neural networks, when trained on image classification data sets, tend to exhibit bias towards texture information, which is the high-frequency element present in images. By contrast, models that are adversarially trained primarily emphasize the importance of low-frequency information, leading to improved robustness~\cite{zhang2019interpreting}. To delve further into this phenomenon, several studies have experimented with adjusting the constraints placed on either low or high-frequency signals in the loss regularization term~\cite{bernhard2021impact}, or incorporating perturbations of variable frequencies during adversarial training~\cite{maiya2021frequency}. There are also several attempts~\cite{zhang2019adversarial,huang2022frequency} aimed at exploiting the above mentioned frequency properties to build more robust models. However, they have primarily been limited to applying low-pass filtering directly to clean or adversarial inputs. Regrettably, this methodology is associated with an irreversible loss of high-frequency information within the image and thus causes a marked decrease in accuracy across clean samples~(e.g., 4-5\% clean accuracy drop compared to standard AT methods)~\cite{huang2022frequency}. Furthermore, it is imperative to adapt the parameters of the low-pass filter to better suit different datasets that embody distinct frequency characteristics.

In order to solve the aforementioned problems and to adopt a new perspective to study the frequency characteristics of robust models, we propose a plug-and-play module called Frequency Preference Control Module~(FPCM) to reconfigure the low-frequency and high-frequency components of the intermediate features learned within the model, which can be simply cooperated with any adversarial training framework and further boost model's robustness. While previous methods require special tuning of their introduced hyperparameters to adapt to datasets with various frequency characteristics, our module can be co-optimized with the model in adversarial training with no need for any fine-grained hyperparameter adjustments, which also allows it to adapt and then utilize frequency features exhibited in different stages~(i.e., layers) of a model. Our empirical study reveals that our approach demonstrates adaptability across different model architectures and datasets and continuously drives robust accuracy improvement. Moreover, with FPCM, we can pioneer the study of the frequency characteristics of intermediate-level feature representations and take a closer look at how the frequency bias of a robust model would impact its robustness.

In a nutshell, our contribution can be folded into following aspects:
\begin{enumerate}
    \item We proposed the Frequency Preference Control Module that can be effortlessly incorporated into any AT approach and further improve the model robustness. 
    \item Leveraging our proposed module, we show how the frequency of feature representation would impact the final robustness and adversarial training process.
\end{enumerate}






\section{Related Work}
\subsection{Adversarial Attack}
Szegedy~\etal~\cite{szegedy2013intriguing} found the vulnerability of DNNs against adversarial examples and proposed L-BGFS based attack. Subsequently, Goodfellow~\etal~\cite{goodfellow2014explaining} argued that the main reason for the vulnerability of neural networks to adversarial perturbations is their linear nature,  and used Fast Sign Gradient Method~(FGSM) to generate adversarial examples efficiently. Kurakin~\etal~\cite{kurakin2016adversarial} extended FGSM to a more effective iteration-based attack as Basic Iterative Method~(BIM). Subsequently, by adding random initialization and extending to more iterative steps, Madry~\etal~\cite{madry2017towards} propose PGD as a universal first-order adversary. Boundary-based attack DeepFool~\cite{moosavi2016deepfool} and optimization-based attacks like C\&W~\cite{carlini2017towards}.

In addition to the thriving white-box attack methods based on the assumption of model transparency, researchers are also especially interested in black-box attacks, where the structure and parameters of the DNN are unknown to the attackers.~(i.e., only a few queries or training data are available). There are two types of black-box attacks: query-based and transfer-based. To produce adversarial instances, query-based algorithms estimate the gradient information via queries~\cite{brendel2017decision,ilyas2018black,andriushchenko2020square}. While transfer-based attacks~\cite{dong2019evading,xie2019improving,salzmann2021learning} employ white-box attacks on a local surrogate model to produce transferable adversarial perturbations.

AutoAttack~\cite{croce2020reliable} is a combination of four complementary attack methods~(i.e., APGD-CE, APGD-DLR, FAB~\cite{croce2020minimally}, and Square Attack~\cite{andriushchenko2020square}, which has become a very popular adversarial robustness benchmark today.

\subsection{Adversarial Training as a Defense}
Adversarial training is one of the notable defense methods as an empirical defense, which is also adopted in this paper to build robust models. And its basic idea can be expressed as a min-max optimization problem:
$$
\theta^* = \mathop{\arg\min}\limits_{\theta} \mathbb{E}_{(x,y)\in \mathcal{X}}\left[\mathop{\max}\limits_{\delta \in [-\epsilon, \epsilon]}\ell(f_{\theta}(x+\delta); y)\right]
$$
where the $f_\theta$ is a DNN parameterized by $\theta$, $\mathcal{X}$ stands for the training dataset and $\ell$ represents the loss function. And perturbations $\delta$ are bounded into the $\epsilon$-ball. 

To solve the inner maximization problem, projected gradient descent (PGD)~\cite{madry2017towards} is a prevailing and effective method to generate perturbations. Many promising methods then sprang up. Some of them provide more efficient loss function designs~\cite{zhang2019theoretically,wang2020improving}, some methods design stronger regularization methods~\cite{wu2020adversarial,qin2019adversarial}, and some focus on solving the problem of the excessive computational overhead of adversarial training~\cite{shafahi2019adversarial,wong2020fast}.

\subsection{Frequency Analysis of ViTs and CNNs}
In engineering applications, the vast majority of Fourier transform applications use the discrete Fourier transform (DFT), where the Fourier transform takes a discrete form in both the time and frequency domains. Applying DFT to a flattened image signal $x$ is equal to left multiplying a DFT matrix, the rows of which are the Fourier basis $f_k = [e^{2\pi j(k-1)\cdot 0},...,e^{2\pi j(k-1)\cdot(n-1)}]/\sqrt{n}\in \mathbb{R}^{n}$, where $k$ represents the $k$th row of the DFT matrix and $j$ represents the imaginary unit. The Fast Fourier Transform~(FFT) is usually used in practical applications to efficiently compute the DFT.

Recent work~\cite{wang2022anti,park2022vision} studied and revealed the low-pass characteristic of basic ViT blocks. ViT tends to reduce high-frequency signals in the feature map and thus almost only the DC component is preserved with the model going deeper. Additionally, MSAs spatially smoothen feature maps with self-attention weights, acting like a natural denoising module for better generalization and robustness. With rigid theoretical analysis, Wang \etal~\cite{wang2022anti} try to avoid over-smoothing by giving higher weights to high-frequency signals in deep ViTs. On the contrary, convolutional networks go the opposite way in terms of frequency characteristics and amplify high-frequency signals. As per the aforementioned findings and the underlying connection between Vit's low-frequency characteristics and its robustness, it is promising to supplement the low-frequency features that convolutional networks are missing.

\begin{figure}[t]
    \centering
    \includegraphics[width=0.8\linewidth]{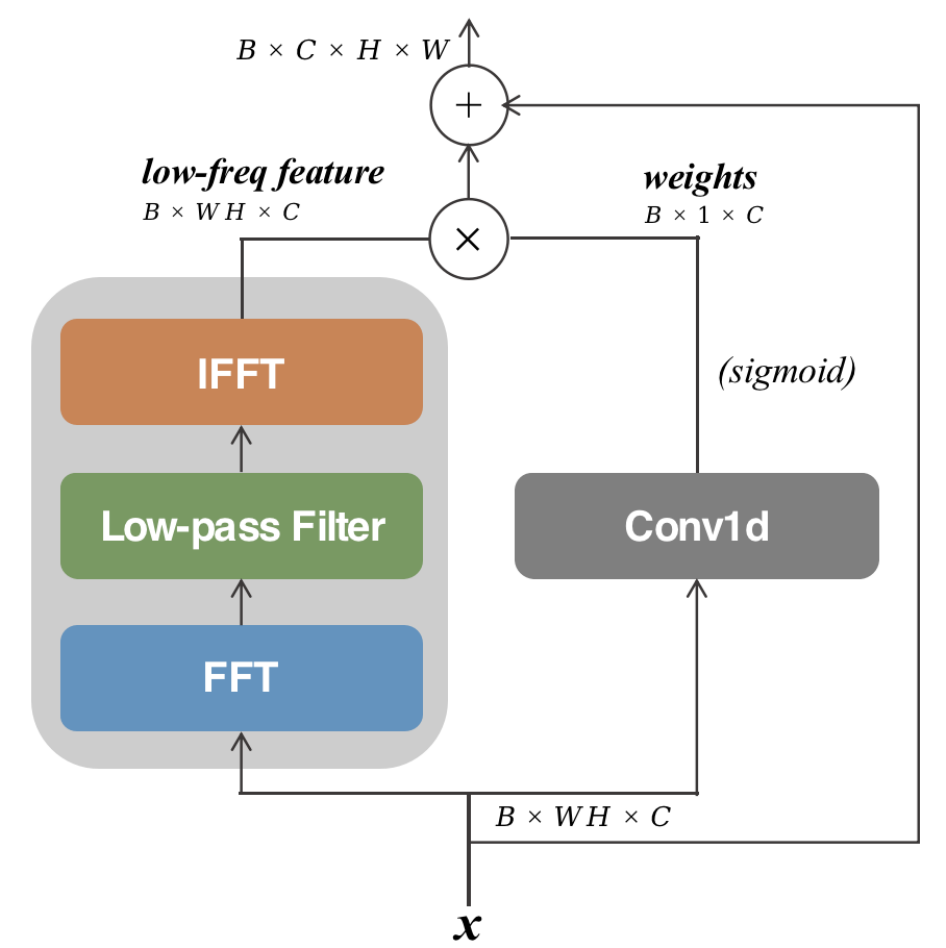}
    \caption{The structure of FPCM}
    \label{fig:LEEM}
\end{figure}

\section{Methodology}

\subsection{Frequency Preference Control Module~(FPCM)}

\begin{figure*}[t]
    \centering
    \includegraphics[width=0.95\linewidth]{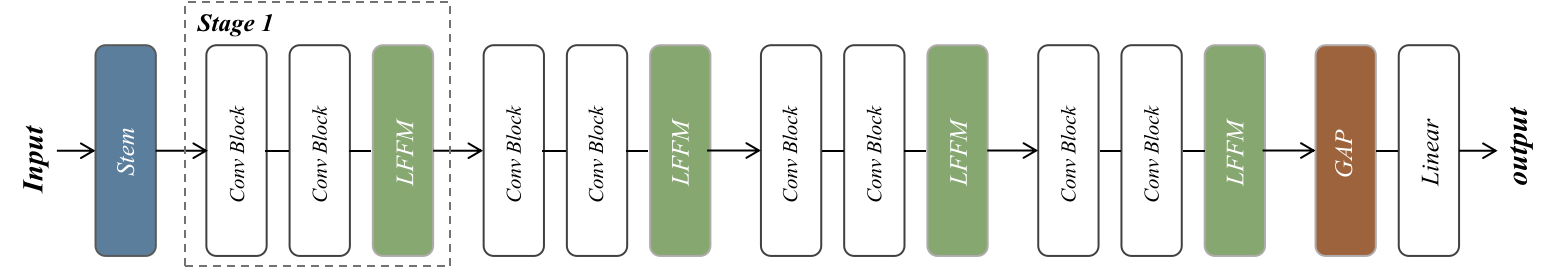}
    \caption{Detailed architecture for our model altered from ResNet18.}
    \label{fig:overview}
\end{figure*}

\begin{figure}[t]
    \centering
    \includegraphics[width=0.8\linewidth]{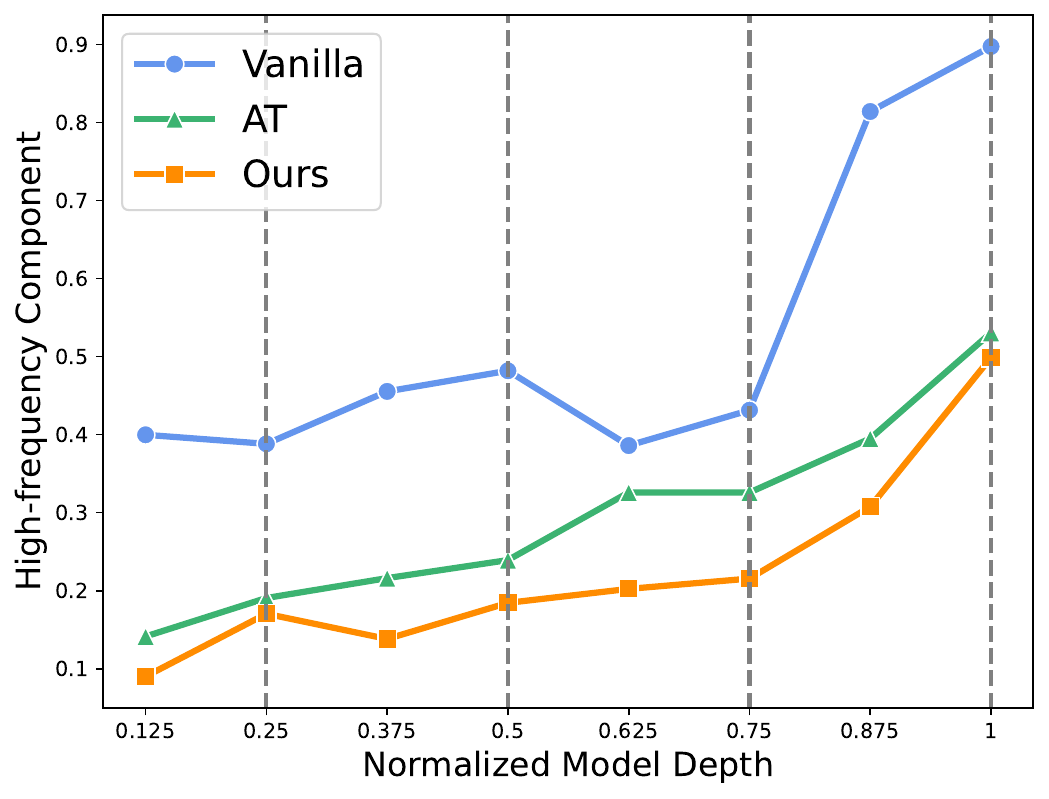}
    \caption{Frobenius norm of the every layers' output high-frequency signals. Convolutional layers increase high-frequency components in feature maps. The vertical grey dashed line represents the last layer at every stage.}
    \label{fig:freq_depth}
\end{figure}

We first introduce the proposed FPCM~(shown in Fig.\ref{fig:LEEM}), which reconfigures low- and high-frequency components in features. Given an input $x$, we transform it into a frequency domain via DFT. We then suppress the high-frequency signal with an exponential low-pass filter, followed by the iDFT transformation to restore spatial domain features. 

Formally, denote $x\in \mathbb{R}^{C \times H\times W}$ as the input feature, and $\mathcal{F} \in \mathbb{C}^{HW \times C}$ as its frequency representation. We have:
$$
\mathcal{F}(u,v) = \sum_{a=0}^{H-1}\sum_{b=0}^{W-1} x_{a,b} e^{-j 2\pi (\frac{v}{H}a + \frac{u}{W}b)}
$$
where $H$ and $W$ represent the height and width of the input feature respectively, and $j$ is the imaginary unit. In practice, the transformation from the discrete spatial domain to frequency domain can be done by FFT with an optimized computational complexity of $O(HWlog(HW))$. Subsequently, the filtered frequency representation $\hat{\mathcal{F}}$ is formulated by:
$$
\hat{\mathcal{F}}(u,v) = F(u,v)H(u,v)
$$
where $H(u,v)$ denotes a low-pass filter in frequency domain. In our work, we mainly study the Gaussian low-pass filter~(GLPF), which is a common choice in the field of image processing, causing no ringing effect:
$$
H(u,v) = e^{-(\frac{D(u,v)}{2D_{0}})^2} = e^{-(\frac{D(u,v)}{2(HW * \beta)})^2}
$$
where $D(u,v)$ represents the distance from the pixel $(u,v)$ to the centre of the two-dimensional spectrum, and $D_0$ is the cutoff frequency. For any feature map with height $H$ and width $W$, the total sequence length of the corresponding Fourier spectrum would be $HW$. Correspondingly, We set $\beta$ as a hyper-parameter controlling the cut-off frequency. Having obtained the filtered frequency, we then apply the Fourier inverse transform to restore the feature map with only the low-frequency components $\hat{x}$, and derive the final re-weighted low-frequency features from the element-wise production of $\hat{x}$ and the inter-channel weights $\alpha \in \mathcal{R}^C$ learned from the original input:
$$
\alpha = 0.5 * sigmoid(W_{\alpha}^{T}x) + 0.5
$$
by default, we push the value interval of $\alpha$ to $[0.5, 1]$ to make the model focus more on the low-frequency features, inspired mainly by previous studies on robust learning frequency characteristics and success attempts~\cite{park2022vision}. Overall, the FPCM can be formulated as the weighted summation of low-frequency and high-frequency features:
$$
FPCM(x) = \alpha \hat{x} + (1-\alpha)(x - \hat{x}) 
$$
we also studied FPCM with fixed weights $\alpha$, where our method will become completely parameter-free. The design of FPCM follows the following principles:

\begin{itemize}
    \item Efficiency: the extra parameter and computational overhead it introduces are marginal compared to standard DNNs
    \item Differentiability: FFT and the subsequent low-pass filtering are essentially linear transformations that do not mask the gradient, which allows joint optimization with DNNs.
    \item Adaptability: learnable weights capture the inter-channel frequency characteristics and bring the possibility of adapting the frequency differences of the output feature maps between different stages~(layers) of a model, various models, and datasets.
\end{itemize}

\subsection{Harmonizing FPCM with Regular CNNs}
In light of prior research demonstrating that convolutional layers typically increase feature map variance, with the variance tending to increase with depth and peaking at the ends of each stage, our investigation focuses on analyzing the feature map frequency characteristics of both a vanilla ResNet18 model and an adversarially trained one. We present the results in Fig.\ref{fig:freq_depth}, which illustrate that high-frequency feature components tend to aggregate as the model goes deeper. The learned high-frequency features of each stage peak at the stage's final layer, as indicated by a greater quantity of high-frequency components at the layer marked by the dashed line when compared to the previous layer. Our findings also demonstrate that, in comparison to the vanilla model, the adversarially trained model prioritizes low-frequency features, with an overall reduction in the number of high-frequency components, consistent with the findings outlined in \cite{huang2022frequency} regarding the smoothing of convolutional kernels by adversarial training. Based on these observations, we hypothesize that incorporating FPCMs closer to the end of each stage can effectively reduce high-frequency signals and further enhance model robustness. To this end, we present an overview of our meta-model in Fig.\ref{fig:overview}, utilizing the ResNet18 architecture as an example. We insert FPCM "painlessly" without destroying the original model architecture, which also works for other kinds of models like WideResNets.

\subsection{Frequency Principle of Deep Learning}
Recently, Xu \etal \cite{xu2022overview} studied the F-principle of deep learning both empirically and theoretically, revealing that DNNs first model the low-frequency features in data. The F-Principle is a salient and intuitive notion, as it aligns with the principles governing human visual perception. Specifically, when humans encounter unfamiliar stimuli, they tend to process and retain coarse, high-level information about its shape and structure prior to encoding more detailed features. Empirically, the weight loss landscape is a widely used indicator to characterize the standard generalization gap in a standard training scenario. For robustness learning, it is also indicated in previous work that a flatter loss landscape leads to higher test robustness as long as the training is sufficient. Also in line with the frequency principle, early stopping can improve the generalization ability of DNNs in practice.

Taking these mentioned aspects into consideration, we endeavor to enhance the concordance between the learning process of the original model and FPCM by modifying the cutoff frequency. Intuitively, we adjust the cutoff frequency linearly as follows:
$$
D_{0_t} = \frac{S}{2} + \frac{t*(\frac{S}{8} - \frac{S}{2})}{T}
$$
where $t$ is the current model training epoch, $T$ represents the total number of training epochs, and $S$ is the length of the frequency spectrum. Therefore, in the process, we gradually decrease the cutoff frequency from $S/2$ to $S/8$. The same setting is used for the perturbations generated during training, but the cutoff frequency is fixed to $S/8$ during evaluation. The impetus behind our approach is to expedite the process of convergence while simultaneously enabling the model to attend to information with complete frequency during the initial training phase. However, we aim to employ more low-frequency signals in the terminal stage, which we believe is to be beneficial for robust learning.


\begin{table*}[htbp]
	\centering
 \renewcommand{\tabcolsep}{10pt}
	\begin{tabular}{c | cccccc}
		\toprule  
        Method & Clean & PGD-10 & PGD-20 & PGD-50 & C\&W & AutoAttack \\
        \midrule
        \midrule
        PGD-AT &85.32   &55.17   &54.28  &53.98   &53.38    &51.43\\
        TRADES &84.72 &56.75 &56.10 &55.9 &53.87 &52.30\\
        MART &84.17 &58.98 &58.56 &58.06 &54.58 &51.10\\
        FAT & 87.97 &50.31 &49.86 &48.79 &48.65 &47.48\\
        GAIRAT &86.30 &59.64 &58.91 &58.74 &45.57 &40.30 \\
        AWP & 85.57& 58.92 &58.13 &57.92 &56.03 &53.90 \\
        LBGAT & \textbf{88.22} &56.25 &54.66 &54.30 &54.29 &52.23 \\
        LAS-AT & 86.23 &57.64 &56.49 &56.12 &55.73 &53.58 \\
        \midrule
        \midrule
        Ours-AT &85.97 &55.35   &54.45   &54.24   &53.93 & 51.87\\
        Ours-TRADES &84.63   &57.91   &57.32   &57.26   &53.93 & 52.88\\
        Ours-AWP & 85.15 & 59.76 & 59.01 & 58.87 & 56.73 & 55.18\\
        Ours-AWP-LAS & 88.10 & \textbf{60.97} & \textbf{59.67} &\textbf{59.31} &\textbf{57.38} & \textbf{55.47}\\
		\bottomrule  
	\end{tabular}
\caption{Test robustness (\%) on the CIFAR-10 database using WRN34-10. Number in bold indicates the best.}
\label{tab:cifar10-wrn34}
\end{table*}

\begin{table*}[htbp]
	\centering
	\renewcommand{\tabcolsep}{10pt}
	\begin{tabular}{c | cccccc}
		\toprule  
        Method & Clean & PGD-10 & PGD-20 & PGD-50 & C\&W & AutoAttack \\
        \midrule
        \midrule
        PGD-AT &85.13   &51.13   &50.17   &49.82   &48.54& 47.42 \\
        TRADES &84.43   &53.71      &52.53       &52.28      &50.81& 49.55 \\
        FAT &\textbf{87.72}& 47.46& 46.69& 46.33 &49.66 &43.14\\
        GAIRAT &83.40 &55.81 &54.76 & 54.21 & 38.81 & 31.25\\
        AWP &82.66 & 55.64 & 54.95 & 54.93 & 50.92 & 50.21\\
        LAS-AT &82.40 & 54.31 & 53.45 & 53.10 & 50.76 & 49.91\\
        \midrule
        \midrule
        Ours-AT & 82.90 & 53.35 & 52.50 & 52.29 & 51.01 & 49.26\\
        Ours-TRADES &82.12 & 55.41 & 54.75 & 54.71 & 50.93 & 49.93\\
        Ours-AWP &82.84 & \textbf{56.49} & \textbf{55.94} & \textbf{55.86} & \textbf{51.81} & \textbf{51.28}\\
		\bottomrule  
	\end{tabular}
 \caption{Test robustness (\%) on the CIFAR-10 database using ResNet18. Number in bold indicates the best.}
 \label{tab:cifar10-resnet18}
\end{table*}
\section{Experiment}
In this section, we conduct a set of experiments to establish the efficacy of the proposed approach. Subsequently, we investigate the effects of parameters such as cutoff frequency and weights for low-frequency features on both clean and robust accuracy. Further, we undertake an in-depth analysis of the frequency domain characteristics of adversarial training, and distinguish our method from other frequency-based techniques.
\subsection{Evaluation Setup}
\paragraph{Training} We conduct experiments on CIFAR10 \& CIFAR100~\cite{krizhevsky2009learning} mainly with ResNet18~\cite{he2016deep} and WRN-34-10~\cite{zagoruyko2016wide}, both of which are the mainstream models for robustness evaluation. We train ResNet18 for 120 epochs with the learning rate initialized at 0.01 and reduced to 1e-3 and 1e-4 at the 84th~($0.7\times120$) and 108th~($0.9\times120$) epoch. For WRN-34-10, the learning rate is initially started at 0.1 and decays to one-tenth and at the 42nd and 54th epoch respectively throughout the 60-epoch training process. The optimizer is SGD with momentum set to 0.9 and the weight decay factor of 3.5e-3 and 5e-4 for ResNet18 and WRN-34-10 respectively. We train and evaluate the baseline counterpart (PGD-AT~\cite{madry2017towards} and TRADES~\cite{zhang2019theoretically}) of our methods with the same training and evaluation settings to make a fair comparison. For the method we built based on AWP~\cite{wu2020adversarial}, we trained according to the settings proposed in their work, extending the number of training epochs to 200 for ResNet and 150 for WRN for performing sufficient training, given its stronger regularization properties. We also compare our method~(Ours-AT, Ours-TRADES, and Ours-AWP) with the following baselines: 1)\ MART~\cite{wang2020improving}, 2)\ FAT~\cite{zhang2020attacks}, 3)\ GAIRAT~\cite{zhang2020geometry}, 4)\ LBGAT~\cite{cui2021learnable} and 5)\ LAS-AT~\cite{jia2022adversarial}.

\noindent
\textbf{Evaluation} We employ several prevailing attack methods to evaluate the robustness of trained models, including PGD, C\&W$_\infty$, and AutoAttack. Unless otherwise specified, all attack methods are performed with perturbation budget $\epsilon=8/255$ under $\ell_{\infty}$. PGD is equipped with random-start with attack step size set to $\alpha=2/255$. The total step of C\&W is 50. We report the averaged result of 5 independent runs where clean and robust accuracy are used as the evaluation metrics. We also leverage the Weighted Robust Accuracy (W-Robust)~\cite{gurel2021knowledge} to help measure the trade-off between clean and robust accuracy, it is defined as follows:
$$
\mathcal{A}_f = \pi_{D_{nat}}P_{D_{nat}}[f(x)=y] + \pi_{D_{adv}}P_{D_{adv}}[f(x)=y]
$$
where $\mathcal{A}_f$ are the accuracy of a model $f$ on $x$ drawn from either the clean distribution $D_{nat}$ or adversarial distribution $D_{adv}$. We adopt $\pi_{D_{nat}} = \pi_{D_{adv}} = 0.5$ by default, treating clean and robust accuracy equally for comprehensive evaluation. 

\subsection{Robustness Comparison}

Our method is a plug-and-play component with no interventions in adversarial training. Therefore, it can be easily combined with other AT methods like TRADES which provides optimized loss design, and LAS which introduces an automatic strategy for generating adversaries.

\noindent
\textbf{Comparisons on CIFAR-10 dataset.}
The results on CIFAR-10 of WRN-34-10 and ResNet18 are listed in Table.\ref{tab:cifar10-wrn34} and Table.~\ref{tab:cifar10-resnet18}. The three proposed models all exceed their counterpart baseline models under most of the attack scenarios. Considering the trade-off between clean accuracy and robustness, our method makes notable improvement on robust accuracy with no significant sacrifice of generalization ability on clean samples. For example, with WRN-34-10 as the experimented model, we boost the performance under PGD-50 attack by 0.95\% and AutoAttack by 1.28\% upon the powerful AWP method. We can further push up the robustness under PGD-50 by 0.66\% and under AA by 0.29\% with LAS to a generate better strategy for perturbation, which also shows the compatibility of our method. In terms of our ResNet18 model based on PGD-AT, we further improve the robustness under PGD-50 by 2.47\%, C\&W by 2.47\%, and AA by 1.84\%. For both ResNet18 and WRN-34-10, our method based on AWP achieves the best performance under all attacks. 

In order to provide additional evidence of the efficacy of our proposed methodology, we endeavored to create a model using WRN-70-16 architecture and compared its performance against state-of-the-art robust models without additional training data. The results are presented in Table.~\ref{tab:sota_compare}. Our approach, build upon AWP, was found to yield greater robust accuracy under AA. Moreover, the performance enhancement yielded by our method is on par or outperforms that observed with the potent LAS technique. Notably, we also can build a more robust model based on LAS-AWP.

\begin{table}[t]
	\centering
	\small
	\begin{tabular}{c | cccc}
		\toprule  
        Method & Clean & PGD-20 & C\&W & AA \\
        \midrule
        \midrule
        PGD-AT &\textbf{86.6}  &62.5    &60.4  & 59.3  \\
        Ours-AT &86.5 & \textbf{63.0}  &\textbf{61.5} & \textbf{60.5}\\
        \midrule
        \midrule
        TRADES &\textbf{83.7}  &64.8  &62.1 &61.9\\
        Ours-TRADES &83.5 & \textbf{65.3} & \textbf{62.8} & \textbf{62.4}\\
		\bottomrule  
	\end{tabular}
 \caption{Test robustness (\%) on the Imagenette database using ResNet18. Number in bold indicates the best.}
 \label{imagenette}
\end{table}

\begin{table}[t]
	\centering
	\small
	\begin{tabular}{c | cccc}
		\toprule  
         Method & Clean & PGD-20 & C\&W &AA \\
        \midrule
        \midrule
        PGD-AT &57.75 & 27.18 &23.41 &22.37\\
        TRADES &56.75 & 28.05 &24.17 &22.89\\
                                 Ours-AT &\textbf{58.79} & 28.75 & \textbf{26.60} & \textbf{25.19}\\
                                 
                                 Ours-TRADES &56.89 & \textbf{30.44} & 26.33  & 25.13\\
        \midrule
        \midrule
        PGD-AT &60.21 & 30.78 &29.50 &27.39\\
        TRADES &58.43 & 29.23 &27.05 &25.78\\
                                  Ours-AT &\textbf{60.22} & \textbf{31.95} &\textbf{30.14} &\textbf{27.81}\\
                                  
                                  Ours-TRADES &57.75 & 31.88 &27.93 &26.98\\
		\bottomrule  
	\end{tabular}
 \caption{Test robustness (\%) on the CIFAR-100 database using ResNet18~(top column) and WRN-34-10~(bottom column). Number in bold indicates the best.}
 \label{CIFAR-100}
\end{table}

\begin{table}[t]
    \centering
    \small
    \begin{tabular}{c|c c}
    \toprule
         Method & Clean & AutoAttack \\
    \hline
         Gowal~\etal~\cite{gowal2020uncovering}& 85.29 & 57.20 \\
         LAS-AWP~\cite{jia2022adversarial} & 85.66 & 57.61\\
         Ours-AWP & \textbf{85.94} & \textbf{57.62}\\
    \bottomrule
    \end{tabular}
    \caption{Clean and robust accuracy (\%) on the CIFAR-10 with WRN-70-16}
    \label{tab:sota_compare}
\end{table}

\begin{table*}[t]
  \begin{center}
    \begin{tabular}{c | c c c c c c c | c c}  
    \toprule
     \multirow{2}{*}{Metrics} & \multicolumn{7}{c}{Fixed weights} & \multicolumn{2}{|c}{Learnable}\\
     & 0 & 0.1 & 0.25 & 0.5 & 0.75 & 0.9 & 1 & Conv1d & MLP \\
     \hline
     Clean  &-& 80.01 & 81.25 &81.86& 82.34 & 81.97 & 82.27 & 81.87 & \textbf{82.83}\\
     PGD-20 &-& 53.19 & 53.81 &53.91& 53.76 & 54.23 & 54.16 & \textbf{54.75} & 54.10\\
     W-Robust &-&66.60 & 67.53 & 67.94 & 68.05 & 68.10 & 68.22& 68.31 & \textbf{68.47}\\
     \hline
     Param & \multicolumn{7}{c}{ - } & 1,360 & 0.35M \\
    \bottomrule
    \end{tabular}
    \caption{Clean and robust accuracy on CIFAR10 dataset of our ResNet18 model with various weights for low-frequency features.}
    \label{tab:alpha_ablation}
  \end{center}
\end{table*}

\noindent
\textbf{Comparisons on more complicated datasets.}
To demonstrate the generalization ability of our methods, we conduct experiments on CIFAR-100 which contains more categories than CIFAR-10 and Imagenette which is of higher resolution~(160*160). The Imagenette dataset is a subset of ImageNet, consisting of ten easily classifiable categories. The results are given in Table~\ref{imagenette} and Table~\ref{CIFAR-100}. Our models all exceed their baselines in terms of robustness. Specifically, on the imagenette database, we improve performance under AA by 1.2\% compared to PGD-AT and 0.5\% in comparison to TRADES. On CIFAR-100 with ResNet18 as the targeted model, the robustness under PGD-20 and AA increases by 2.39\% and 2.24\% respectively. All the above mentioned numbers reveal that our method can be easily combined with AT frameworks and further improve their performance steadily across different models and datasets.

\subsection{Ablation Studies}
We utilize two hyper-parameters that merit further investigation, where $\alpha$ governs the trade-off between low- and high-frequency components and $\beta$ regulates the cutoff frequency of the low-pass filter.
\vspace{-0.3cm}
\subsubsection{Cutoff frequency}
Results of various settings for cutoff frequency are listed in Table.~\ref{tab:cutoff}. With the weights $\alpha\in \mathcal{R}^{C}$ staying learnable, we first try to set fixed $\beta$ for the whole training and evaluation process. A smaller beta equates to less high-frequency information being retained. It is consistent with the intuition that the clean accuracy drops with $\beta$ set to be smaller, while the robust accuracy goes the opposite way. This suggests that convolutional networks rely on high-frequency texture information for better classification performance on clean samples. In contrast, robust models tend to rely more on low-frequency signals to achieve greater robustness, which could partially explain why the clean accuracy of robust models is generally lower than that of vanilla models. We also tried with learnable parameters $\hat{\beta} \in \mathcal{R}^C$, initialized as an all-ones vector, to realize adaptive cutoff frequency for different channels. The possible explanation for the absence of progress in contrast to fixed values might be associated with its redundancy with learnable frequency weights $\alpha$. Upon examining the inner numerical values acquired, an intriguing phenomenon is revealed whereby the estimated coefficient $\hat{\beta}$ exhibits a tendency to diminish in deeper stages, which may indicate that less high frequency is needed in deeper layers for robust learning. Lastly, our proposed linear adjustment of the cutoff frequency during training yields the highest W-Robust.

\begin{table}[t]
    \centering
    \small
    \begin{tabular}{c|c|c|ccc}
    \toprule
         &\multirow{2}{*}{Linear} & \multirow{2}{*}{Learnable} & \multicolumn{3}{c}{Fixed}  \\
         & & & 0.5 &0.25 &0.125 \\
    \hline
         Clean &82.12 & 81.89 & \textbf{82.56} & 81.15 & 80.32\\
         PGD-20 &54.75 & 54.85 & 52.34 & 55.12 & \textbf{55.34}\\
         AA & 49.96& 49.03 & 49.08 & 49.66 & \textbf{50.13}\\
    \hline
         W-Robust &\textbf{66.04} &65.46 &65.82 & 65.41 & 65.23\\
    \bottomrule
    \end{tabular}
    \caption{Clean and robust accuracy of our method with the cutoff frequency factor $\beta$ varying.}
    \label{tab:cutoff}
\end{table}

\begin{table}[t]
    \centering
    \begin{tabular}{c|c c c c}
    \toprule
         Stages & Max & Min & Mean & Var\\
    \midrule
         stage1 & 0.9648 & 0.5393 & 0.6879 & 0.0064\\
         stage2 & 0.9777 & 0.5354 & 0.7024 & 0.0103\\
         stage3 & 0.8047 & 0.7403 & 0.7697 & 0.0001\\
    \bottomrule
    \end{tabular}
    \caption{Some statistics about the learned weights $\alpha$ at different stages of a WRN-34-10 model.}
    \label{tab:alpha_statistic}
\end{table}

\vspace{-0.3cm}
\subsubsection{Re-weighing of low frequency components}

The weights balancing between low- and high-frequency feature components play an important role in robust learning. Therefore, we conduct experiments with $\alpha$ ranging from 0 to 1, as shown in Table.\ref{tab:alpha_ablation}. Due to the fact that a majority of information is in DC components, the model cannot properly converge with $\alpha$ set to 0. Altering $\alpha$ to a value greater than 0.5, i.e., prioritizing low-frequency parts, leads to enhanced accuracy concerning both robustness and clarity, thereby elevating W-Robust by approximately 0.37\%-1.79\%. Incorporation with learnable weights elevates the performance further by 0.39\% on W-Robsut. While we use only one convolutional layer with minimal parameters to develop relationships among the different channels in our initial setup, we further study the FPCM regarding more parameters and non-linear operations~(indicated as \textit{MLP}), but the improvement is marginal~(i.e., +0.03\% W-Robust). 

We proceeded to examine the learned weights and computed relevant statistics as illustrated in Table.\ref{tab:alpha_statistic}. As per our previous observation, we noticed that deeper stages generally had higher weights assigned to low-frequency components compared to earlier stages. For instance, the average $\alpha$ in stage3 was observed to be 0.7697 while it was merely 0.6879 in stage1. It should be highlighted that the weight value in stage3 did not drop below 0.74 emphasizing the significance of low-frequency signals aiding robust learning in deeper layers. From an alternative perspective, we observe that robust models aim to counteract the issue highlighted in Fig. \ref{fig:freq_depth}, where high-frequency signals gradually accumulate in CNNs. By analyzing the statistics of various channels and samples, we have observed that the sample-wise variance of the learned weights is typically small (i.e., usually lower than 1e-4). This implies that the FPCM approach has been tailored to alter the intrinsic frequency characteristics of the robust model, instead of being data-specific.

\subsection{Discussions}

\begin{figure}
    \centering
    \includegraphics[width=0.8\linewidth]{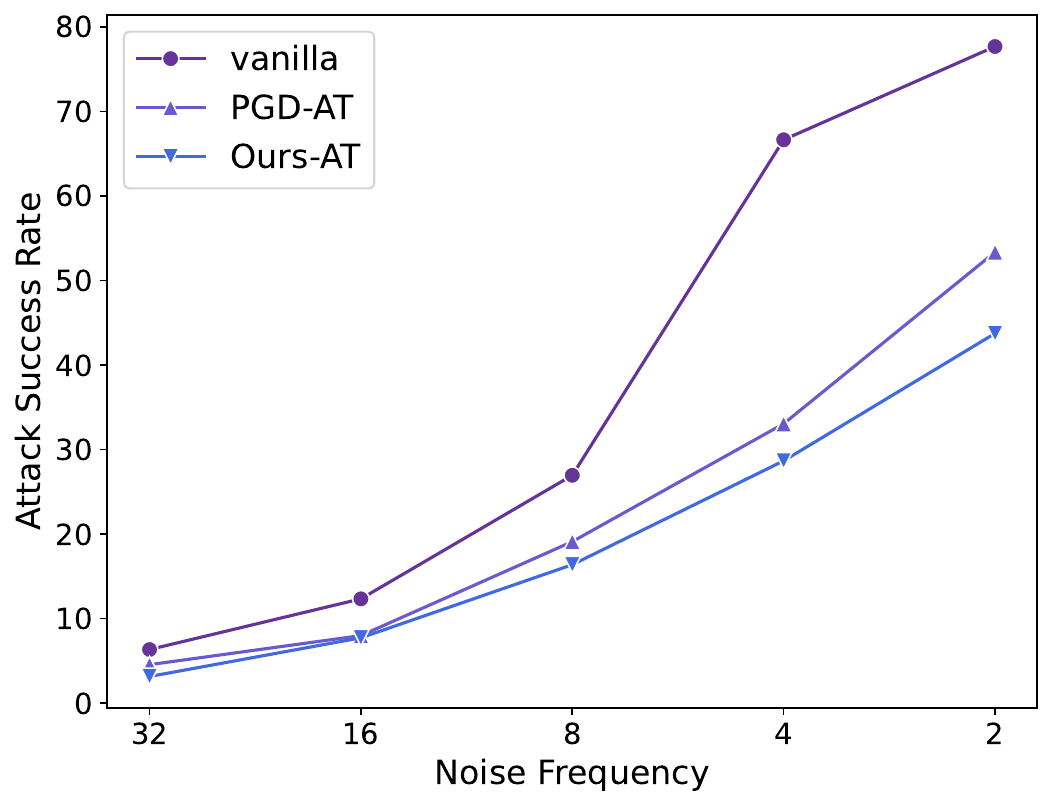}
    \caption{The attack success rate (\%) of noise with respect to its frequency. The numbers on the x-axis indicates the cut-off frequency factor $1/\beta$.}
    \label{robsutness_frequency}
\end{figure}

\subsubsection{Frequency-bias of robust learning}

Previous work has proposed that convolutional layers have high-pass characteristics that cause vanilla ResNet models to be more sensitive to high-frequency signals and thus more vulnerable to high-frequency noise. Our experiment in Fig.\ref{fig:freq_depth} does verify this claim, finding that the high-frequency signal becomes richer as the network gets deeper. In this section, we delve deeper into the frequency characteristics of adversarially trained models. 

We first study the robustness under noise containing different proportions of high-frequency components, as shown in Fig.~\ref{robsutness_frequency}. We still adopt the low-pass filter formulation introduced in Sec.3.1 to conduct this experiment. Therefore, a smaller $\beta$ indicates less high-frequency signal contained in the perturbation. We measure the accuracy drop with frequency-based random noise and the perturbed data can be formulated by:
$x_{noise} = x_{0} + \mathcal{F}^{-1}(\mathcal{F}(\delta) \odot H)$, where $x_{0}$ is the clean data, $\mathcal{F}(\cdot)$ and $\mathcal{F}^{-1}(\cdot)$ are Fourier transform and inverse Fourier transform, $\delta$ is random noise, and $H$ denotes the filter. For the vanilla model, the slope of the line graph increases gradually with more high-frequency components, indicating its vulnerability to high-frequency noise. While similar patterns can be seen in PGD-AT, PGD-AT significantly reduces the model vulnerability against high-frequency noises, which indicates AT models naturally focus more on the low-frequency features in data. Furthermore, by explicitly letting the model focus on the low-frequency components, we further improve the robustness of the adversarially trained model to high-frequency noise, resulting in an overall more robust model. In contrast, the robustness to very low-frequency signals does not improve considerably from vanilla to PGD-AT and Ours-AT.

\begin{figure}[t]
\centering
\subfigure[Robust Acc]{
\begin{minipage}[t]{0.5\linewidth}
\centering
\includegraphics[width=\linewidth]{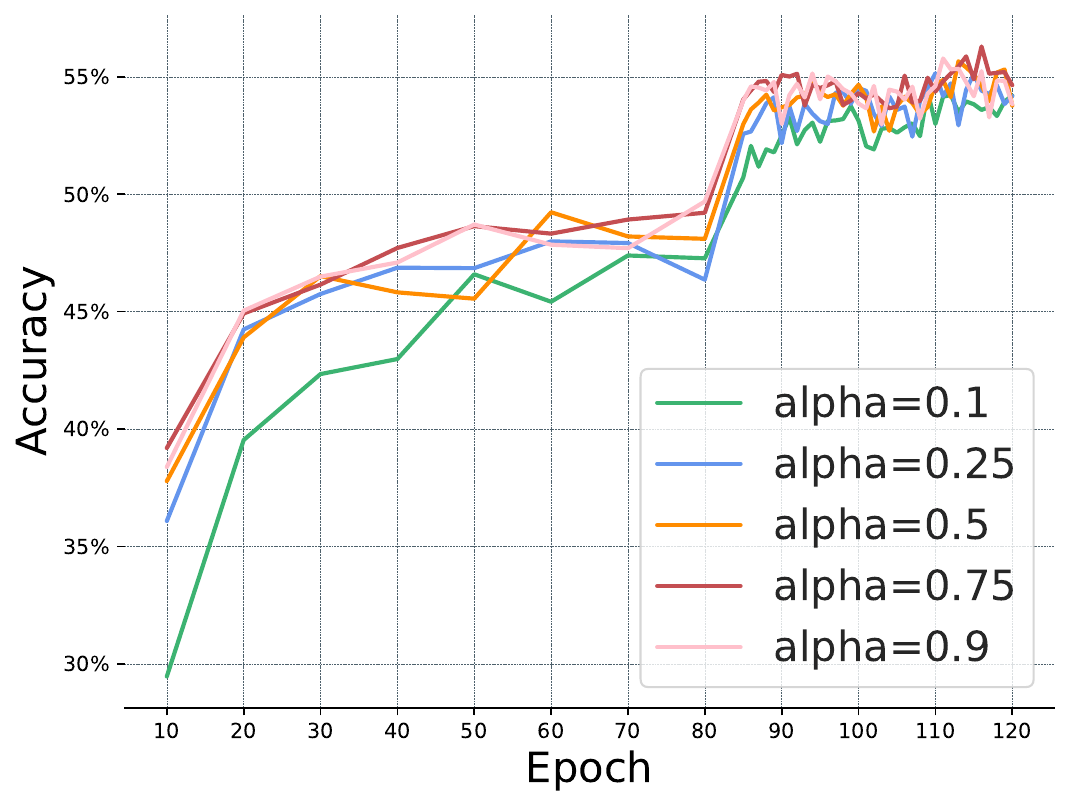}
\end{minipage}%
}%
\subfigure[Loss]{
\begin{minipage}[t]{0.5\linewidth}
\centering
\includegraphics[width=\linewidth]{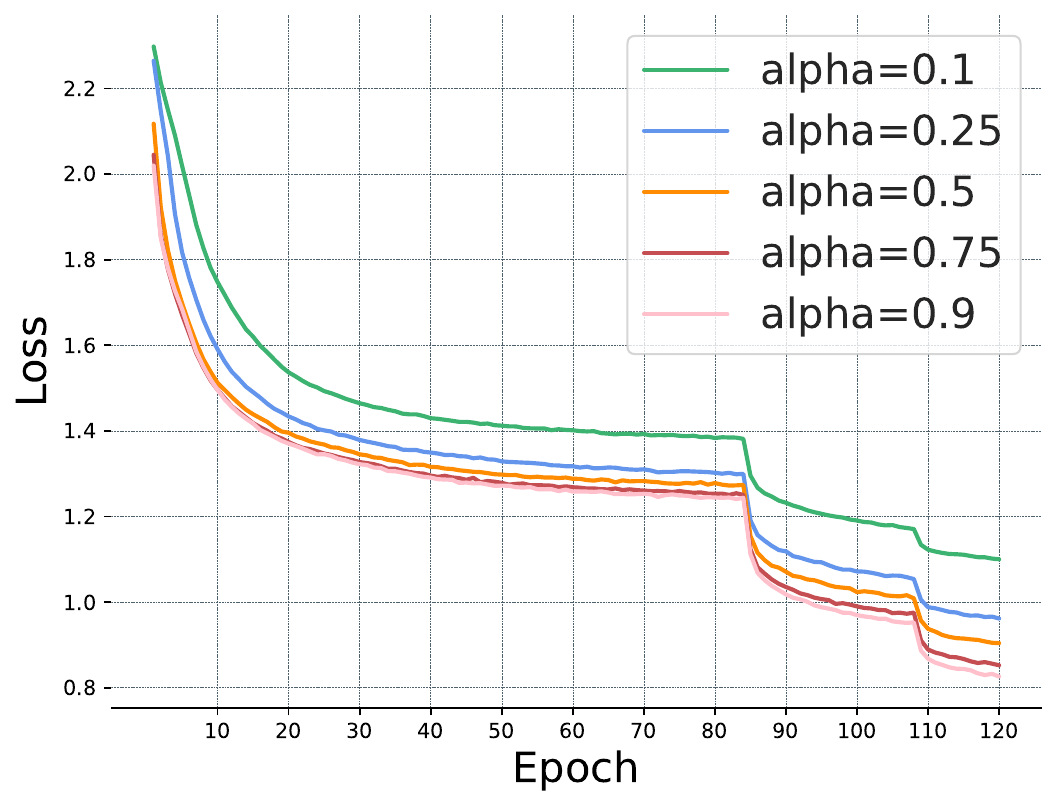}
\end{minipage}%
}%
\caption{Robust accuracy and loss landscape during the training process. For the first 80 epochs, we only keep the accuracy for every 10 epoch for simplicity. Here we use $\beta=0.125$.}
\vspace{-0.4cm}
\label{fig:loss}
\end{figure}

We then take a look at the learning curve of our method with fixed weights $\alpha$ for low-frequency features. Results are plotted in Fig.\ref{fig:loss}. With $\alpha$ set to 0.1, the robust accuracy is lower than other settings during the whole training process. The plot of the loss landscape may partially explain this phenomenon as it is not sufficiently trained on the training data with impaired low-frequency signals. As the $\alpha$ continues to rise, the loss curve moves further down, indicating low-frequency signals can accelerate the robust learning process. We believe that all of the above phenomena illustrate the dominance of low-frequency features in robust learning.

\vspace{-0.3cm}
\subsubsection{Differences \& advantages over other frequency-based methods}

Zhang \etal~\cite{zhang2019adversarial} proposed to suppress high-frequency components in the input image to perform adversarial defense. Following the idea of applying low-pass filtering (LPF) to the natural and adversarial inputs, Huang \etal~\cite{huang2022frequency} additionally introduce a frequency regularization (FR) term, which aims at bridging the frequency domain distance~(measured by $L_{1}$) of clean and adversarial data, by applying DFT on the output logits. They further tried to smooth the convolutional kernels by leveraging Stochastic Weight Averaging (SWA)~\cite{izmailov2018averaging}. Bernard \etal~\cite{bernhard2021impact} also employs regularization on the frequency of output logits with original and low-pass filtered images as the input. Orientating real-world super resolution models, Yue \etal~\cite{yue2021robust} adopts frequency masks to block out high-frequency components in input that possibly contain harmful perturbations in a stochastic manner. Since FR~\cite{huang2022frequency} is the state-of-the-art method derived from the frequency perspective, we make further comparison with it in clean and robust accuracy, and the results are listed in Table. \ref{tab:FR}. It can be seen that our method not only has obvious advantages under AA~(i.e., 0.83\% accuracy improvement), but retains greater generalization on clean data~(+4.06\%). The significant decrease in the accuracy of FR on clean samples may be due to its direct filtering of the input images, which leads to irretrievable signal components that are instrumental for the model to correctly classify clean samples.

In a nutshell, we are the first to focus on the frequency of intermediate features rather than that of the input clean data or adversary.
Our method enhances the robustness of models by explicitly modifying the frequency bias of robust models, as opposed to relying on stronger regularization techniques.

\begin{table}[t!]
    \centering
    \small
    \begin{tabular}{c|c c c c c}
    \toprule
         Method & Clean & PGD-20 & C\&W & AA & W-Robust \\
         \hline
         FR & 80.59 & 59.49 &54.33 &52.06 & 66.33\\
         FR-SWA & 81.09 &\textbf{60.12} &56.14 &54.35 & 67.72\\
         Ours & \textbf{85.15} & 59.01 & \textbf{56.73} & \textbf{55.18} & \textbf{70.17}\\
    \bottomrule
    \end{tabular}
    \caption{Clean and robust accuracy of WRN-34-10 on CIFAR-10 dataset. W-Robust is the average of \textit{Clean} and \textit{AA}.}
    \label{tab:FR}
\end{table}

\vspace{-0.15cm}
\section{Conclusion}
We proposed the Frequency Preference Control Module (FPCM) to adaptively reconfigure low- and high-frequency components in the intermediate feature representations. Based on our finding that high-frequency signals tend to aggregate as the model deepens, we designed a heuristic scheme to insert FPCM after each stage of the regular model to promote harmonious operation among them. Empirical studies demonstrate that our method can be easily incorporated into any adversarial training framework and further boost its performance with minimal computation overhead. Furthermore, by leveraging FPCM, we conducted experiments that shed light on the frequency characteristics of robust models, revealing insights regarding the impact of low-frequency bias on their robustness.

\section{Acknowledgement}
The work is supported in part by HK RIF (R7030-22), HK ITF (GHP/169/20SZ), National Key R\&D Program of China (2022ZD0160200), the Huawei Flagship Research Grants in 2021 and 2023, the HKU-SCF FinTech Academy R\&D Funding Schemes in 2021 and 2022, and the Shanghai Artificial Intelligence Laboratory (Heming Cui is a courtesy researcher in this lab).

{\small
\bibliographystyle{ieee_fullname}
\bibliography{egbib}
}

\end{document}